\documentclass{article}
\pdfoutput=1

\usepackage{arxiv}

\usepackage[utf8]{inputenc} 
\usepackage[T1]{fontenc}    
\usepackage{hyperref}       
\usepackage{url}            
\usepackage{booktabs}       
\usepackage{amsfonts}       
\usepackage{nicefrac}       
\usepackage{microtype}      
\usepackage{float}

\usepackage{amsmath}
\usepackage{amssymb}

\usepackage{indentfirst}
\usepackage{multirow, multicol}
\usepackage{caption}
\usepackage{algorithm}
\usepackage{algpseudocode}

\usepackage{todonotes}

\usepackage{tikz}
\usetikzlibrary{decorations.pathmorphing}

\title{Using dynamical quantization to perform split attempts in online tree regressors}

\author{
  Saulo Martiello Mastelini\\
  Institute of Mathematics and Computer Sciences\\
  University of São Paulo\\
  São Carlos, BR \\
  \texttt{mastelini@usp.br} \\
  \And
  André C. Ponce de Leon Ferreira de Carvalho\\
  Institute of Mathematics and Computer Sciences\\
  University of São Paulo\\
  São Carlos, BR \\
  \texttt{andre@icmc.usp.br} \\
}

\begin{document}
\maketitle

\begin{abstract}

A central aspect of online decision tree solutions is evaluating the incoming data and enabling model growth.
For such, trees much deal with different kinds of input features and partition them to learn from the data.
Numerical features are no exception, and they pose additional challenges compared to other kinds of features, as there is no trivial strategy to choose the best point to make a split decision.
The problem is even more challenging in regression tasks because both the features and the target are continuous.
Typical online solutions evaluate and store all the points monitored between split attempts, which goes against the constraints posed in real-time applications.
In this paper, we introduce the Quantization Observer (QO), a simple yet effective hashing-based algorithm to monitor and evaluate split point candidates in numerical features for online tree regressors.
QO can be easily integrated into incremental decision trees, such as Hoeffding Trees, and it has a monitoring cost of $O(1)$ per instance and sub-linear cost to evaluate split candidates.
Previous solutions had a $O(\log n)$ cost per insertion (in the best case) and a linear cost to evaluate split points.
Our extensive experimental setup highlights QO's effectiveness in providing accurate split point suggestions while spending much less memory and processing time than its competitors.

\end{abstract}


\section{Introduction and Background}

In the last few years, given the growing production of data, data stream mining has received increased attention.
These data might come from different sources, such as the internet, Internet of Things (IoT) devices, sensors, and many other options.
Data streams are potentially unbounded and might change through time.
These characteristics gave risen to multiple unsupervised and supervised learning algorithms to process this flow of endless data.
In the context of supervised data stream mining, online or incremental Decision Tree (DT) algorithms have been a frequent option among researchers and practitioners~\cite{krawczyk2017ensemble}.

DTs are powerful in the sense of inducing accurate predictors, which can be even improved when combined in ensembles~\cite{ke2017lightgbm,gomes2017adaptive}.
They are also flexible learners since no assumptions are made about the data distribution.
DTs can be inherently interpreted, which is highly desirable in Machine Learning (ML) applications~\cite{molnar2019}.
Besides, DT induction algorithms can be adapted to deal with non-stationary distributions that might be subjected to concept drifts (CD)~\cite{bifet2009adaptive, manapragada2018extremely}.

However, online DT (ODT) models face additional constraints when compared with their traditional batch counterparts.
First, while the data is unbounded, the computational resources are limited.
For this reason, ODTs can neither store instances indefinitely nor process them multiple times~\cite{gama2010knowledge, bifet2018machine}.
Typical solutions process each incoming datum once, which is then discarded.
Besides, from the start, the trees must be able to predict new instances and be updated anytime.
Hence, ODT models should be maximally accurate, whereas keeping the memory and processing time usage minimal.

Multiple families of theoretical ODT algorithms have been proposed over the years.
The most popular of them, the Hoeffding Tree (HT), relies on Hoeffding's inequality~\cite{hoeffding1963probability} to decide when an incremental model has gathered enough information to expand itself.
Popular realizations of HTs include the Very Fast Decision Tree (VFDT), for classification
~\cite{domingos2000mining}, the Fast Incremental Regression/Model Tree (FIRT/FIMT), for regression~\cite{ikonomovska2011learning}, and the incremental Structured Output Prediction Tree (iSOUP-Tree), for structured output tasks~\cite{osojnik2018tree}.
Other ODT were created using the same theoretical framework~\cite{pfahringer2007new, ikonomovska2015online}, and also decision rule systems~\cite{almeida2013adaptive}.
Among the ODTs that do not fit in the HT framework, we can mention the McDiarmid Trees~\cite{rutkowski2012decision} and the Stochastic Gradient Trees~\cite{gouk2019stochastic}.

Apart from their core differences, all ODTs share a common property: they monitor input features and perform split attempts.
ODTs must process the stream's features and store statistics relating each input to the target value as data continuously arrives.
These statistics differ for classification or regression tasks~\cite{domingos2000mining, ikonomovska2011learning}.
The stored statistics will enable the models to evaluate split candidates for each feature and decide upon the best feature (and split/cut point) to grow the tree structure.
For such, ODTs rely on a class of algorithms named Attribute Observers (AO).
ODTs carry one AO per feature in each one of their leaves.

ODTs can efficiently deal with categorical attributes since split enabling statistics can be directly maintained for each category.
Numerical attributes, on the other hand, do not have explicit partitions and might not
be trivially manipulated to calculate split points~\cite{pfahringer2008handling, ikonomovska2011learning}.
Usually, batch DT rely on sorting operations to evaluate split candidates.
At each node, the tree has to sort the numerical input values and evaluate every available binary split decision.

Regardless of their domain (batch or online tasks), DT induction algorithms typically use similar heuristics to calculate the merit of a split.
Classification trees often use entropy~\cite{quinlan2014c4} or gini impurity~\cite{breiman1984classification} as metrics to evaluate splits.
Regression trees typically rely on error-reducing measures, such as the squared error (in practice equivalent to variance reduction)~\cite{breiman1984classification} or the absolute error.

In the online setting, the cost of performing sorting operations is prohibitive.
Thus, less computationally expensive alternatives have been proposed to overcome this limitation~\cite{pfahringer2008handling, gouk2019stochastic}.
Rather than sorting the attributes, typical ODT solutions use data structures or attribute distribution estimators to keep the input values and their statistics sorted with reduced costs.
For example, early ODTs for classification and most of the current versions for regression use a binary search tree (BST) to store input values and target statistics.
This structure is named Extended Binary Search Tree (E-BST) since it stores both input values and target statistics in its nodes to evaluate split candidates.
While classification tree's nodes store class counts~\cite{domingos2000mining}, in regression tasks, the nodes maintain variables used for online mean and variance calculation procedures~\cite{ikonomovska2011learning,osojnik2018tree}.

Although fairly efficient, E-BST still has an insertion cost of $O(\log n)$, in the best case, and a memory cost of $O(n)$, where $n$ is the number of stored instances.
Moreover, when evaluating split candidates, an in-order traversal of the BST is needed to account for all monitored input values.
Such traversal, or split point query, has a cost of $O(n)$.
Another limitation of E-BST  is that only partial information is available for evaluation at the split time, in contrast to batch DT solutions.
Consequently, the split chosen for an input attribute after observing $n$ instances might not be the same had the tree monitored $n+k$ instances.
If extrapolation can be applied, the obtained split points might be improved.

More efficient algorithms have been investigated for classification ODTs~\cite{pfahringer2008handling}.
In these tasks, the target attribute has well-defined partitions (categories), hence,
more effective data structures or estimation strategies can be applied to the AOs.
One of the solutions broadly applied in these tasks is the usage of histograms to monitor the class counts for different discretized intervals~\cite{pfahringer2008handling}.
Online histogram solutions typically have a $O(\log m)$ cost per insertion and $O(m)$ of memory cost and query cost, where $m$ represents the number of histogram bins.
Popular tree gradient boosting ensemble frameworks, such as LightGBM~\cite{ke2017lightgbm}, also use histograms to discretize numerical attributes, even though they process the data in the traditional batch approach.

Another popular solution is to approximate the probability density distribution of each class in a numerical attribute using Gaussian distributions~\cite{pfahringer2008handling,gama2010knowledge}.
These distributions can be easily constructed in an online fashion and only require estimating the sample mean and variance.
Thus, one can approximate the probabilities involved in the class impurity measures.
Even better, this strategy has a cost of $O(1)$ for adding a new observation.
The Gaussian Naive Bayes algorithm uses a similar approximation to deal with numerical features~\cite{flach2012machine}.

Unfortunately, in regression tasks, more than counts are required to calculate the dispersion measures for split candidate evaluation.
The Variance Reduction (VR) strategy applied to evaluate how promising is each split point involves keeping incremental estimation of the sample mean and variance.
These estimators must be kept for each split candidate lying in the hyper-rectangle defined by a path from the tree root to a leaf.
The VR-guided tree growth can also be seen as a greedy Mean Square Error (MSE) minimization process~\cite{breiman1984classification}.
In this case, the optimization aims to reduce the MSE by making splits that create tree leaves whose points are maximally close to their centroid, i.e., the mean of the points in that region.

Therefore, incremental tree regressors must be able to calculate variances for each partition created by dividing the input space in axis-aligned splits of the form $x \le c$ (left branch) and $x > c$ (right branch).
In the previous expressions, $x$ is one of the input features, and $c$ is the split threshold (or cut value).

Most of the existing regression ODTs still rely on the E-BST structure, whose disadvantages were previously pointed out.
Besides, the algorithm commonly used in the E-BST for calculating variances is known to be unstable and produce inaccurate results~\cite{finch2009incremental,knuth2014art}.
Consequently, all the current regression ODTs are prone to intensive memory usage and processing time, as well as yielding inaccurate results due to the numerical instability of the incremental variance estimators.

Ideally, we would like to obtain AO algorithms for regression whose cost for adding new examples and querying split candidates is $O(1)$.
However, if we can devise an accurate solution with constant insertion cost and sub-linear query costs, this is already on par with the most advanced AO solutions for ODT classifiers.

This paper introduces the Quantizer Observer (QO), a dynamical quantization algorithm to handle numerical splits in ODT regressors.
QO stores the same kind of targets' statistics as those monitored by E-BST.
However, QO has a $O(1)$ cost per insertion of elements and a memory cost of $O(n')$, where $n' \ll n$.
Despite being much faster than E-BST, QO can still produce split candidates with similar discriminating capability.
To assess its performance, QO is experimentally compared with the traditional E-BST and a variation of E-BST that truncates the input values before their insertion in the BST, named Truncated E-BST (TE-BST).
TE-BST aims to reduce the memory usage and processing time of E-BST.

To circumvent the problem of inaccurate incremental variance estimation, we extended the formulae proposed by Chan et al.~\cite{chan1982updating} to handle distributed variance estimation.
The formulas in \cite{chan1982updating} are, in turn, extensions from the well-known Welford's algorithm~\cite{knuth2014art}, which improves the naive incremental variance estimation currently used in E-BST.
The expressions proposed in \cite{chan1982updating} enable summing partial estimates of the variance.
We extend them by also enabling subtracting partial estimates of variance from each other.
Hence, all the AOs for regression compared in this work adopt these enhanced and robust incremental variance estimators.

We benchmark the different AOs using an extensive synthetic data setup and account for insertion, storage, and query costs.
Here, we focus on the AOs rather than on the actual tree models.
This focus allows us to isolate the splitting procedure from other aspects of the ODTs,
such as tree traversal and how the models compute predictions.
We vary the sample characteristics, size, and noise levels to simulate different situations the AOs might face when working in the trees.
According to the experimental results, QO reduced, with statistical significance, 
the memory costs and processing time when compared with the existing AOs for regression.
The experimental results also show that QO can produce split points similar to those provided by E-BST.

The remaining of this work is organized as follows: Section~\ref{sec_problem} formalizes the split point search in regression ODTs.
Section~\ref{sec_mean_variance} presents the robust incremental mean and variance estimators that replace the previous used unstable estimators.
Next, Section~\ref{sec_qo} presents our proposed AO, QO.
We detail our evaluation setup in Section~\ref{sec_setup} and discuss the obtained results in Section~\ref{sec_results}.
Finally, we present our final considerations and possible directions for future research in Section~\ref{sec_final_considerations}.

\section{Problem definition}\label{sec_problem}

Suppose an infinite stream $S = \left\{(\mathbf{x}, y)_t\right\}_{t=0}^\infty$, where each object at time $t$, $(\mathbf{x}, y)$, is composed of numerical input attributes $x \in \mathbf{x}$, and a scalar target attribute $y$.
Regression ODT induction algorithms work by creating binary partitions in the numerical features.
These partitions take place at a specific attribute value $x=c$.
Hence, trees grow by creating branches from a decision node $d$ to leaf nodes $l_{-}$ and $l_+$, which are defined by the tests $x \le c$ and $x > c$, respectively.
To guarantee that the resulting models will be accurate, the tree learning algorithms have to determine the best $(x, c)$ combination, among all $x \in \mathbf{x}$.

As previously discussed, DT regressors typically minimize the MSE of the target value of points in a leaf node compared to their mean value, which is also referred to as centroid or prototype point~\cite{kocev2013tree}.
This strategy is equivalent to minimizing the variance of the $y$ values belonging to each leaf.
For this reason, when performing a split attempt, DT regressors aim at choosing the partition candidate that maximally reduces the variance of $y$, here simply referred to as $s^2$.

The resulting heuristic to guide tree growth, called Variance Reduction (VR), is defined in Equation~\ref{eq_vr}.

\begin{equation}
    \text{VR}_{(d, \left\{l_-, l_+\right\})} = s^2(d) + \frac{|l_-|}{|d|}s^2(l_-) + \frac{|l_+|}{|d|}s^2(l_+)
    \label{eq_vr}
\end{equation}

In the last expression of this equation, the notation $|.|$ represents the number of instances lying in the tree node inside the brackets.
From Equation~\ref{eq_vr}, it is clear that trees must be able to calculate the variance of the target variable in their nodes.
This operation is trivial in batch regression DT induction algorithms since all data is available beforehand.
In online applications, however, the algorithms must estimate the variance incrementally and at any time for each partition induced by a given realization of $(x,c)$.
AOs, such as E-BST, are used for this end.

Each node in an E-BST represents one of the observed values $x_v$ of the monitored feature $x$.
New observations are added as new leaves in the E-BST, and the order in which the $x$ values are inserted impact how balanced the BST becomes.
Only the nodes accessed when a new instance is sorted down the E-BST have their statistics updated.
Nodes store target statistics accounting for all $x$ observations that are smaller than or equal to their $x_v$.

Originally, E-BST was designed to operate using the so-called naive incremental variance estimator~\cite{knuth2014art}.
Hence, each E-BST node stores $\sum_{x \le x_v} w$, $\sum_{x \le x_v} y$, and $\sum_{x \le x_v} y^2$, respectively, the sum of weights, the target values, and the squared target values.
These properties, although producing inaccurate estimates of the variance and being prone to numerical cancellation~\cite{finch2009incremental}, can be easily merged.
In other words, we can either add or subtract the statistics of two different E-BST nodes and calculate the resulting variance.
Finally, by doing a complete in-order traversal, we can retrieve the statistics necessary to compute the VR value for the partition induced by each $x_v$ in the E-BST.

Next, we discuss how to improve the VR values calculated in the E-BST and any other AO for regression tasks.

\section{Robust Variance calculation}\label{sec_mean_variance}

Calculating the variance of each candidate partition is a central aspect of ODT regressors.
As previously mentioned, the current solutions rely on an incremental algorithm with well-known problems~\cite{finch2009incremental,knuth2014art}.
We start this section by describing the Welford's algorithm, a robust and popular
alternative for calculating variance incrementally~\cite{knuth2014art}.

The Welford's algorithm works by keeping an estimate of the sample mean at the $n$-th instance, $\overline{x}_n$, which is used to update the auxiliary second order statistics $M_{2, n}$.
This auxiliary value is used, in turn, to calculate the variance.
At the beginning of the data monitoring process, we set both $\overline{x}_0$ and $M_{2, 0}$ to zero.
After each new observation $x_n$ arrives, we update the stored statistics, as follows.
The mean estimate update is given by Equation~\ref{eq_mean}.

\begin{equation}
    \overline{x}_n = \overline{x}_{n-1} + \dfrac{x_n - \overline{x}_{n-1}}{n}
    \label{eq_mean}
\end{equation}

The $M_{2, n}$ value is also recursively updated by using Equation~\ref{eq_m2n}.

\begin{equation}
    M_{2,n} = M_{2,n-1} + (x_n - \overline{x}_{n-1})(x_n - \overline{x}_n)
    \label{eq_m2n}
\end{equation}

At any time, one can get an estimate of the monitored sample variance by calculating $s_n^2 = \frac{M_{2,n}}{n-1}$, for $n > 1$.

Chan et al.~\cite{chan1982updating} extended the presented formulae to handle parallel updates.
In other words, by using the new expressions presented next, we can process different parts of the stream separately and then merge the resulting statistics to obtain mean and variance estimates for the whole sample.

In the following equations, for notation simplicity, we do not show the $n$ subscript.
Instead, we add the subscripts $A$ and $B$ to denote two groups of partially monitored data, whose statistics are going to be merged.
We also denote by $AB$ the resulting merged group.
The total number of observed examples can be directly computed as $n_{AB} = n_A + n_B$.
By using $n_{AB}$, we can calculate the total estimate of the mean, using Equation~\ref{eq_mean_ab}.

\begin{equation}
    \overline{x}_{AB} = \dfrac{n_{A}\overline{x}_A + n_{B}\overline{x}_B}{n_{AB}}
    \label{eq_mean_ab}
\end{equation}

Now, we have all the needed tools to estimate the total second order statistic, as defined in Equation~\ref{eq_mean_ab}. In the expression, $\delta = \overline{x}_B - \overline{x}_A$.

\begin{equation}
    M_{2,AB} = M_{2, A} + M_{2,B} + \delta^{2}\dfrac{n_{A}n_{B}}{n_{AB}}
    \label{eq_m2n_ab}
\end{equation}

By doing some simple algebraic manipulations in the equations of Chan et al.~\cite{chan1982updating}, we obtain expressions to govern the subtraction of the incremental statistics.
In other words, one can get the complement of partially monitored sample statistics if they also have complete statistics.
Therefore, we get the same addition and subtraction properties the naive incremental variance calculation algorithm has, but still retain superior accuracy in our estimates.
These more robust incremental mean and variance estimates are going to be used in the AOs for ODT regressors.

We start by the simplest one, the number of objects, which is given by $n_A = n_{AB} - n_B$.
Equation~\ref{eq_mean_a} presents the expression to retrieve $\overline{x}_A$, given $\overline{x}_{AB}$ and $\overline{x}_{B}$.

\begin{equation}
    \overline{x}_A = \dfrac{n_{AB}\overline{x}_{AB} - n_{B}\overline{x}_B}{n_A}
    \label{eq_mean_a}
\end{equation}

Finally, we can get the complement of partial second order statistic by using Equation~\ref{eq_m2n_a}.

\begin{equation}
    M_{2,A} = M_{2, AB} - M_{2, B} - \delta^{2}\dfrac{n_{A}n_{B}}{n_{AB}}
    \label{eq_m2n_a}
\end{equation}

\section{Quantizer Observer}\label{sec_qo}

Our proposal, QO, is inspired by Locality Sensitive Hashing (LSH)~\cite{datar2004locality} algorithms, which are used to approximate nearest neighbor search and also discretize numerical input features.
Our proposal also aims at creating partitions so that similar input values are grouped.
Unlike most LSH algorithms, QO relies on a single hash structure, $H$, to create hash slots (also referred to as buckets) for the discretized features.
Moreover, QO deals with one feature at a time, so there is no need to involve multiple random projections to define hash codes, as in popular LSH solutions.
Instead, we simply define a quantization radius, $r$, to discretize the incoming input feature.
In each of $H$'s slots, QO keeps the sum of $x$'s values, and estimations of the mean and variance of $y$.
QO relies on the equations presented in Section~\ref{sec_mean_variance} to update and combine the target's statistics.

The functioning of QO is straightforward and can be easily incorporated into existing regression tree ODT algorithms.
QO works as follows.
For each $i$-th observation of a feature $x$, we select its corresponding hash code $h$, i.e., the slot it belongs to in $H$, by following the simple projection scheme $h = \left \lfloor \frac{x_i}{r} \right \rfloor$.
If $h$ is not in $H$, we create a new slot to accommodate the incoming data, otherwise, we include the values of $x_i$ and $y_i$ to the existing slot.
Algorithm~\ref{alg_qo_update} illustrates the update procedure of QO, i.e., how AO monitors incoming examples.

\begin{algorithm}[!htb]
    \begin{algorithmic}
        \State{\textbf{Input:}}
        \State{\hspace{1em}$r$: the quantization radius.}
        \State{\hspace{1em}$S_x$: stream containing observations of a numerical feature $x$ and the target $y$.}
        \State{\textbf{Initialization:}}
        \State{\hspace{1em}Let $H$ be an empty hash table.}
        \vspace{1em}
        \For{$x_i, y_i \in S_x$}
            \State{Let $s^2_{y_i}$ be an instance of the robust variance estimator with a single observation $y_i$.}
            \State{Let $h \leftarrow \lfloor\frac{x}{r}\rfloor$ be chosen hash slot.}
            \If{$h \in H$}\Comment{Update statistics and prototype}
                \State{$H[h]_{x} \leftarrow H[h]_{x} + x_i$}
                \State{$H[h]_{s^2} \leftarrow H[h]_{s^2} + s^2_{y_i}$}
            \Else\Comment{Create a new hash slot}
                \State{$H[h]_{x} \leftarrow x_i$}
                \State{$H[h]_{s^2} \leftarrow s^2_{y_i}$}
            \EndIf
        \EndFor
    \end{algorithmic}
    \caption{QO update.}
    \label{alg_qo_update}
\end{algorithm}

Since $h$ is directly proportional to $x_i$, when evaluating split candidates, QO sorts the keys stored in the hash to get an ordered representation of the whole sample.
We retrieve the necessary information to calculate the VR statistic by computing the cumulative sum of the ordered $H$'s elements.
Hence, the split candidate query cost is $O(|H|\log |H|)$, where $|H|$ is the number of slots in $H$.
We experimentally observed that $|H| \ll n$, where $n$ is the total number of observations.

Split points are defined as the average between the prototype attribute values of two consecutive slots in the ordered hash.
We define the prototype as the mean of the $x$ values belonging to a slot.
Other strategies could also be employed, such as interpolating consecutive slots with a regression model.
Nonetheless, to reduce computational costs, we opted for using a simple approach.
The prototype feature value can be easily obtained using the sum of $x$'s values and the number of observations monitored in each slot.
We illustrate the split point query of QO in Algorithm~\ref{alg_qo_query}.

\begin{algorithm}[!htb]
    \begin{algorithmic}
        \State{\textbf{Input:}}
        \State{\hspace{1em}$H$: an existing QO realization of feature $x$.}
        \State{\hspace{1em}$s^2_y$: the variance estimation of the whole $y$ sample.}
        \State{\textbf{Return:}}
        \State{\hspace{1em}$c$: the best found split threshold in $x$.}
        \State{\hspace{1em}$c_{vr}$: the VR value obtained by partitioning $x$ at $c$.}
        \vspace{1em}

        \State{Let $s^2_\text{aux}$, be an auxiliary variance estimator initialized with zero.}
        \State{$c_{vr} \leftarrow \varnothing$}
        \State{$x_\text{aux} \leftarrow 0$}
        \State{$i \leftarrow 0$}
        \For{h in \texttt{sorted}$(H)$}
            \If{$i > 0$}
                \State{Let $x_p \leftarrow \dfrac{H[h]_x}{H[h]_n}$ be the prototype $x$ value in the hash slot.}
                \State{Let $\hat{c} \leftarrow \dfrac{x_\text{aux} + x_p}{2}$} be the candidate split point.
                \State{Let $\hat{c}_{vr}$ be the VR value obtained from $s^2_y$ and $\left\{s^2_\text{aux}, s^2_y - s^2_\text{aux}\right\}$}.
                \If{$c_{vr}=\varnothing$ \textbf{or} $\hat{c}_{vr} > c_{vr}$}
                    \State{$c_{vr} \leftarrow \hat{c}_{vr}$}
                    \State{$c \leftarrow \hat{c}$}
                \EndIf
            \EndIf

            \State{$x_\text{aux} \leftarrow x_p$}
            \State{$s^2_\text{aux} \leftarrow s^2_\text{aux} + H[h]_{s^2}$}
            \State{$i \leftarrow i + 1$}
        \EndFor
    \end{algorithmic}
    \caption{QO split candidate query.}
    \label{alg_qo_query}
\end{algorithm}

\section{Experimental setup description}\label{sec_setup}

This section describes the simulation protocol used in this study to compare QO against E-BST and TE-BST, as well as the evaluation metrics and settings used in the AOs.

\subsection{Simulation protocol}

We evaluated the effectiveness of the AOs when monitoring data samples of varying sizes.
We also considered different types of functions to generate a target attribute value related to the input.
After calculating the target's values, the inputs were also (in some cases) subject to different levels of noise.
In all the cases, the AOs processed the data sequentially, one instance at a time.
After processing the whole sample, the AOs calculated the best split candidate they could provide, given their inner structures.

Table~\ref{tab_simulation_protocol} summarizes the settings used in our data generation protocol.
Note that one of the bimodal distributions used in our experiments is asymmetric, i.e., its modes have different values of standard deviation.
To generate the data samples and the coefficients that define the target, we
repeated the generation protocol ten times, varying at each time the random initialization.
Our final results, obtained by calculating the metrics described in Section~\ref{sec_evaluation_metrics}, were taken by averaging the results obtained in the several executions.

\begin{table}[!htb]
    \renewcommand{\arraystretch}{1.2}
    \centering
    \caption{Description of the simulation protocol utilized in our experiments.}
    \resizebox{\textwidth}{!}{
    \begin{tabular}{p{0.3\textwidth}p{0.7\textwidth}}
        \toprule
        \textbf{Property description} & \textbf{Property value} \\
        \midrule
        \multirow{2}{*}{Sample size} & 50, 100, 200, 400, 500, 750, 1000, 2500, 5000, 7000, 10000, 15000, 25000, 50000, 75000, 100000, 200000, 500000, 1000000 \\ \cline{2-2}
        Sampling distribution & Uniform, Normal, or Bimodal\\ \cline{2-2}
        Target function & Linear (\texttt{lin}) or Cubic (\texttt{cub}) \\ \cline{2-2}
        Amount of noisy instances & 0\% or 10\% \\ \cline{2-2}
        Noise characteristics & $\mathcal{N}(0, 0.1)$ or $\mathcal{N}(0, 0.01)^a$ \\
        \bottomrule
        \midrule
        \textbf{Distribution name} & \textbf{Distribution property} \\
        \midrule
        Normal & $\mathcal{N}(0, 1)$, $\mathcal{N}(0, 0.1)$, $\mathcal{N}(0, 7)$ \\ \cline{2-2}
        Uniform & $[-1, 1]$, $[-0.1, 0.1]$, $[-7, 7]$ \\ \cline{2-2}
        Bimodal$^b$ & $\mathcal{N}(-1, 1)|\mathcal{N}(1, 1)$, $\mathcal{N}(-0.1, 0.1)|\mathcal{N}(0.1, 0.1)$, $\mathcal{N}(-7, 7)|\mathcal{N}(7, 0.1)$\\
        \bottomrule
    \end{tabular}}
    { \small
        \begin{flushleft}
            $^a$Depending on the parameters of the generating distribution. We added normally distributed noise with smaller standard deviation to distributions whose dispersion was also small.\newline
            $^b$We constructed bimodal distributions by sampling from two Normal distributions with equal probability. We use the ``|'' symbol to indicate a concatenation operation.
        \end{flushleft}
    }
    \label{tab_simulation_protocol}
\end{table}

\subsection{Settings used in the Attribute Observers}

E-BST does not have hyperparameters to setup.
TE-BST, on the other hand, was configured to truncate the input values to three decimal places.
Regarding our proposal, we evaluated three variants of QO to evaluate how $r$ impacts on the obtained results, namely:

\begin{itemize}
    \item QO$_{0.01}$: uses a fixed value ($0.01$) to discretize the input features.
    \item QO$_{\sigma \div 2}$: uses the standard deviation of the feature divided by $2$ as the quantization radius.
    \item QO$_{\sigma \div 3}$: uses the standard deviation of the feature divided by $3$ as the quantization radius.
\end{itemize}

Although the standard deviation of the whole monitored sample is not available beforehand in real-world scenarios, we can rely on variance estimates to obtain good approximations.
These approximations will be used when applying QO to ODT algorithms.
Note that the regression ODTs already keep one incremental variance estimator per leaf node to enable split candidate search.
A fixed radius value, such as $0.01$, can be applied at the beginning of the tree construction as a cold-start choice.
Hence, we also add this variant in our experimental setup, which uses a fixed radius regardless of the input data.
QO is going to be integrated into \texttt{river}\footnote{\url{https://riverml.xyz}}, a popular framework for online machine learning.

\subsection{Evaluation metrics}\label{sec_evaluation_metrics}

We selected three performance evaluation metrics to evaluate the AOs.
They measure how accurate were the splits, how much memory the AOs used, and how long the AOs took to process the data and evaluate the best split candidate.

The first measure was the split merit yielded by each AO, i.e., the obtained VR value.
We also calculated the number of elements stored by each AO to estimate their memory usage.
By element, we mean the number of nodes (E-BST and TE-BST) or the number of hash slots (QO).
Since all the AOs store the same set of target statistics, we can rely on the number of elements rather than precisely measuring their actual memory usage.
The time measurements were twofold: we measured the time taken by the AO to monitor the whole sample and the time they spent to produce a split candidate at the end of the stream monitoring.

For all the metrics, the smaller the value, the better. Besides, among all the metrics, the time measurements are the only ones that have a well-defined scale, i.e., they were measured in seconds.

\section{Results and discussion}\label{sec_results}

\begin{figure}[!htb]
    \centering
    \includegraphics[width=\textwidth]{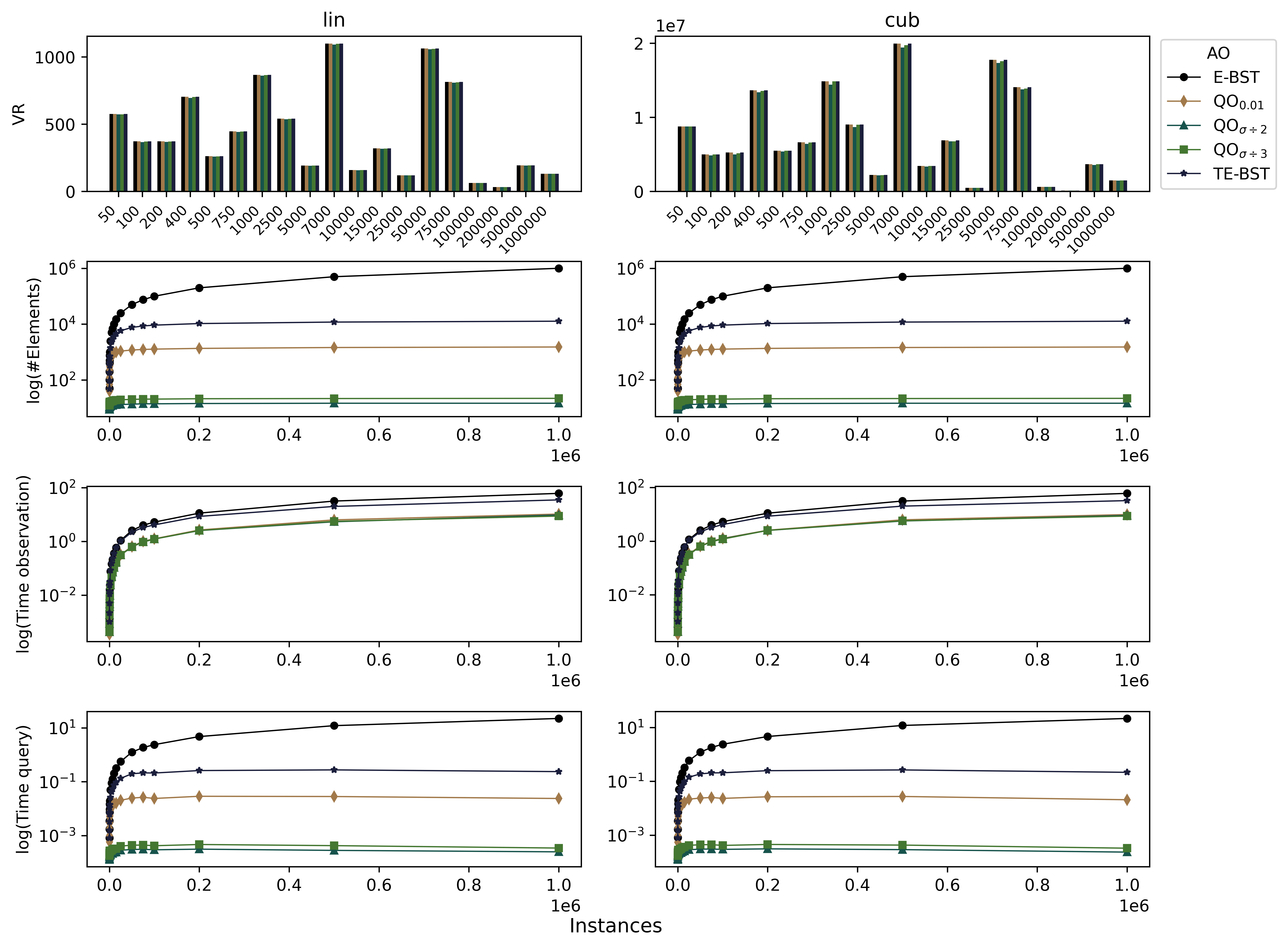}
    \caption{Average results obtained by the compared AOs in the \texttt{lin} and \texttt{cub} tasks. From top to bottom: VR, and the logarithm of the number of stored elements, observation time (in seconds), and query time (in seconds).}
    \label{fig_metrics_results}
\end{figure}

In Figure~\ref{fig_metrics_results}, we summarize, separately for the tasks \texttt{lin} and \texttt{cub}, the average results obtained in the experiments.
We also created separate plots for each data distribution and regression task.
However, looking at the experimental results, we observed that the differences in performance between the AOs followed similar patterns, regardless of the data distribution.
Thus, for simplicity, we only present the charts of the averaged results.

In the next sections, we discuss each evaluation metric separately and refer to specific characteristics observed in Figure~\ref{fig_metrics_results}.
It is worth mentioning that, differently from Figure~\ref{fig_metrics_results}, when performing our in-depth analysis of each metric, we did not average the results between the different data distributions used to generate the synthetic examples.
Instead, we accounted for the results obtained by the AOs, considering each evaluated sample size, data distribution, and regression task.
We relied on Friedman tests and Nemenyi post-hoc tests~\cite{demvsar2006statistical} (with $\alpha=0.05$) to statistically evaluate the statistical significance of the different performances obtained by AOs for each metric.

\subsection{Merit}

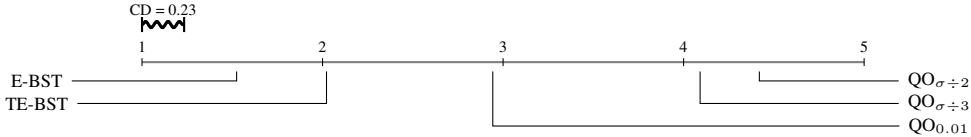
\begin{figure}[!htb]
\centering
\begin{tikzpicture}[xscale=2]
\node (Label) at (1.339943447224415, 0.7){\tiny{CD = 0.23}}; 
\draw[decorate,decoration={snake,amplitude=.4mm,segment length=1.5mm,post length=0mm},very thick, color = black] (1.2,0.5) -- (1.47988689444883,0.5);
\foreach \x in {1.2, 1.47988689444883} \draw[thick,color = black] (\x, 0.4) -- (\x, 0.6);
 
\draw[gray, thick](1.2,0) -- (6.0,0);
\foreach \x in {1.2,2.4,3.6,4.8,6.0} \draw (\x cm,1.5pt) -- (\x cm, -1.5pt);
\node (Label) at (1.2,0.2){\tiny{1}};
\node (Label) at (2.4,0.2){\tiny{2}};
\node (Label) at (3.6,0.2){\tiny{3}};
\node (Label) at (4.8,0.2){\tiny{4}};
\node (Label) at (6.0,0.2){\tiny{5}};
\node (Point) at (1.8289473684210527, 0){};\node (Label) at (0.5,-0.25){\scriptsize{E-BST}}; \draw (Point) |- (Label);
\node (Point) at (2.4263157894736844, 0){};\node (Label) at (0.5,-0.55){\scriptsize{TE-BST}}; \draw (Point) |- (Label);
\node (Point) at (5.303508771929825, 0){};\node (Label) at (6.5,-0.25){\scriptsize{QO$_{\sigma \div 2}$}}; \draw (Point) |- (Label);
\node (Point) at (4.909649122807018, 0){};\node (Label) at (6.5,-0.55){\scriptsize{QO$_{\sigma \div 3}$}}; \draw (Point) |- (Label);
\node (Point) at (3.531578947368421, 0){};\node (Label) at (6.5,-0.85){\scriptsize{QO$_{0.01}$}}; \draw (Point) |- (Label);
\end{tikzpicture}
\caption{Friedman test and Nemenyi post-hoc test when comparing the merit of the splits (VR) generated by the different AO algorithms ($\alpha=0.05$).}
\label{fig_nemenyi_merit}
\end{figure}

As expected, the exhaustive (or near exhaustive) methods presented the highest merit values (VR).
In fact, E-BST and TE-BST consistently surpass the QO variants when it comes to the merit of the obtained splits, as shown in our statistical significance test (Figure~\ref{fig_nemenyi_merit}).
Nonetheless, the actual obtained VR values were very similar, regardless of the AO.
We highlight this fact in the top portion of Figure \ref{fig_metrics_results}.
In these charts, we present the average VR obtained in tasks \texttt{lin} and \texttt{cub}.
It can be seen that the bars representing different AOs' split merits are similar for equal sample size, even considering that VR has a squared operation in its formulation, i.e., an operation that stretches the output range of the heuristic.

\begin{figure}[!htb]
    \centering
    \includegraphics[width=\textwidth]{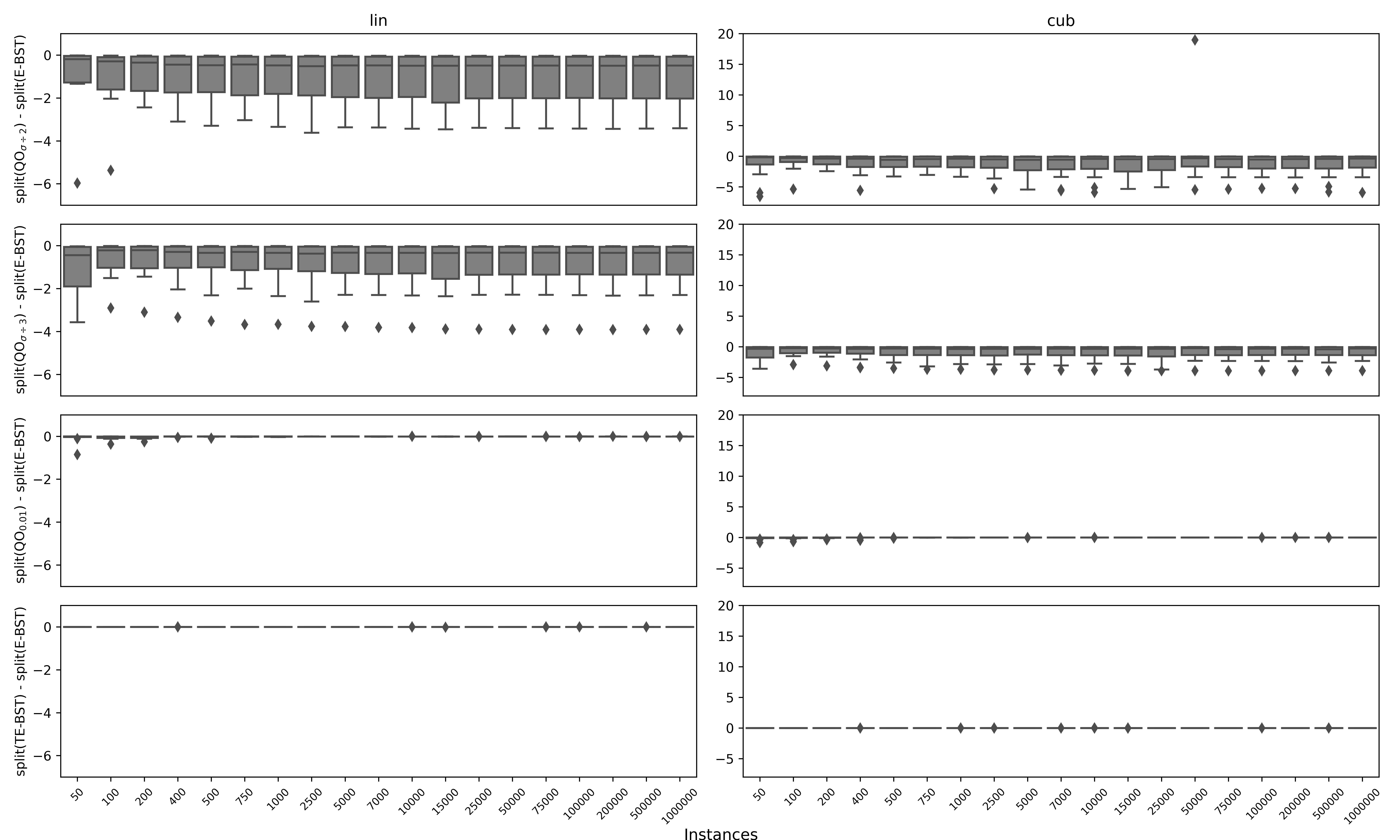}
    \caption{Average differences between the split points found by QO and TE-EBST in comparison with E-BST.}
    \label{fig_boxplot_splits}
\end{figure}

In Figure~\ref{fig_boxplot_splits}, we compare the differences between the split points obtained by TE-BST and QO against the E-BST ones.
We observe that as we decrease QO's quantization radius, its splits become closer to those estimated by E-BST.
The size of this radius is directly correlated with both the obtained merit and the AO size/runtime: the smaller the radius, the higher the merit; similarly, the larger the radius, the smaller the runtime and memory usage.

\subsection{Number of elements}

We start our discussion about the number of stored elements by recalling that the 
amount of memory used by an AO is directly proportional to the number of elements it carries.
In this study, element means a slot (QO) or node (E-BST/TE-BST) carrying a split value and target statistics.

The QO variants obtained the best rankings in this analysis.
The reader might refer to the second row of Figure~\ref{fig_metrics_results} to see how massive this difference was.
The results are presented in log-scale so that we can visually compare the AOs.
Without this visualization trick, the QO results would not be visible in the chart since our proposal used significantly less memory than its competitors.

\begin{figure}[!htb]
\centering
\begin{tikzpicture}[xscale=2]
\node (Label) at (1.339943447224415, 0.7){\tiny{CD = 0.23}}; 
\draw[decorate,decoration={snake,amplitude=.4mm,segment length=1.5mm,post length=0mm},very thick, color = black] (1.2,0.5) -- (1.47988689444883,0.5);
\foreach \x in {1.2, 1.47988689444883} \draw[thick,color = black] (\x, 0.4) -- (\x, 0.6);
 
\draw[gray, thick](1.2,0) -- (6.0,0);
\foreach \x in {1.2,2.4,3.6,4.8,6.0} \draw (\x cm,1.5pt) -- (\x cm, -1.5pt);
\node (Label) at (1.2,0.2){\tiny{1}};
\node (Label) at (2.4,0.2){\tiny{2}};
\node (Label) at (3.6,0.2){\tiny{3}};
\node (Label) at (4.8,0.2){\tiny{4}};
\node (Label) at (6.0,0.2){\tiny{5}};
\node (Point) at (1.2, 0){};\node (Label) at (0.5,-0.25){\scriptsize{QO$_\sigma \div 2$}}; \draw (Point) |- (Label);
\node (Point) at (2.4, 0){};\node (Label) at (0.5,-0.55){\scriptsize{QO$_{\sigma \div 3}$}}; \draw (Point) |- (Label);
\node (Point) at (5.989473684210526, 0){};\node (Label) at (6.5,-0.25){\scriptsize{E-BST}}; \draw (Point) |- (Label);
\node (Point) at (4.803508771929825, 0){};\node (Label) at (6.5,-0.55){\scriptsize{TE-BST}}; \draw (Point) |- (Label);
\node (Point) at (3.6070175438596492, 0){};\node (Label) at (6.5,-0.85){\scriptsize{QO$_{0.01}$}}; \draw (Point) |- (Label);
\end{tikzpicture}
\caption{Friedman test and Nemenyi post-hoc test when comparing the number of elements stored by the different AO algorithms ($\alpha=0.05$).}
\label{fig_nemenyi_size}
\end{figure}
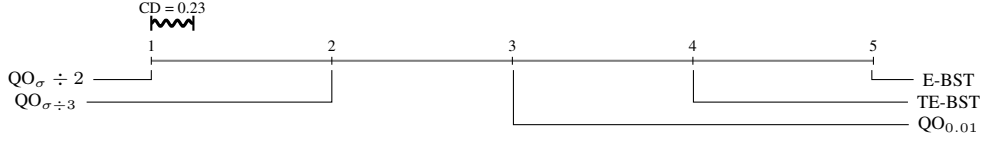

The differences in memory usage are also highlighted in the performed statistical tests, as shown in Figure~\ref{fig_nemenyi_size}.
As expected and shown in the chart, the larger the quantization radius, the more reduced is QO's memory footprint.
In our experimental setup, the fixed radius $r=0.01$ was smaller than the dynamical choices ($\sigma \div 2$ and $\sigma \div 3$).
Hence, QO with a constant quantization radius spent more memory than the other variants.
On the other hand, concerning split merit, the QO$_{0.01}$ was also the most accurate variant.
Users might use smaller proportions of the feature's standard deviation to balance the split merit and the computational costs.
Lastly, as we also expected, TE-BST stored significantly fewer elements than E-BST.

\subsection{Time}

\begin{figure}[!htb]
\centering
\begin{tikzpicture}[xscale=2]
\node (Label) at (1.339943447224415, 0.7){\tiny{CD = 0.23}}; 
\draw[decorate,decoration={snake,amplitude=.4mm,segment length=1.5mm,post length=0mm},very thick, color = black] (1.2,0.5) -- (1.47988689444883,0.5);
\foreach \x in {1.2, 1.47988689444883} \draw[thick,color = black] (\x, 0.4) -- (\x, 0.6);
 
\draw[gray, thick](1.2,0) -- (6.0,0);
\foreach \x in {1.2,2.4,3.6,4.8,6.0} \draw (\x cm,1.5pt) -- (\x cm, -1.5pt);
\node (Label) at (1.2,0.2){\tiny{1}};
\node (Label) at (2.4,0.2){\tiny{2}};
\node (Label) at (3.6,0.2){\tiny{3}};
\node (Label) at (4.8,0.2){\tiny{4}};
\node (Label) at (6.0,0.2){\tiny{5}};
\draw[decorate,decoration={snake,amplitude=.4mm,segment length=1.5mm,post length=0mm},very thick, color = black](2.2868421052631582,-0.25) -- (2.504385964912281,-0.25);
\draw[decorate,decoration={snake,amplitude=.4mm,segment length=1.5mm,post length=0mm},very thick, color = black](5.234210526315789,-0.4) -- (5.55,-0.4);
\node (Point) at (2.336842105263158, 0){};\node (Label) at (0.5,-0.65){\scriptsize{QO$_{\sigma \div 3}$}}; \draw (Point) |- (Label);
\node (Point) at (2.424561403508772, 0){};\node (Label) at (0.5,-0.95){\scriptsize{QO$_{\sigma \div 2}$}}; \draw (Point) |- (Label);
\node (Point) at (5.5, 0){};\node (Label) at (6.5,-0.65){\scriptsize{TE-BST}}; \draw (Point) |- (Label);
\node (Point) at (5.284210526315789, 0){};\node (Label) at (6.5,-0.95){\scriptsize{E-BST}}; \draw (Point) |- (Label);
\node (Point) at (2.454385964912281, 0){};\node (Label) at (6.5,-1.25){\scriptsize{QO$_{0.01}$}}; \draw (Point) |- (Label);
\end{tikzpicture}
\caption{Friedman test and Nemenyi post-hoc test when comparing the time spent by the different AO algorithms to monitor the input data ($\alpha=0.05$).}
\label{fig_nemenyi_time_obs}
\end{figure}
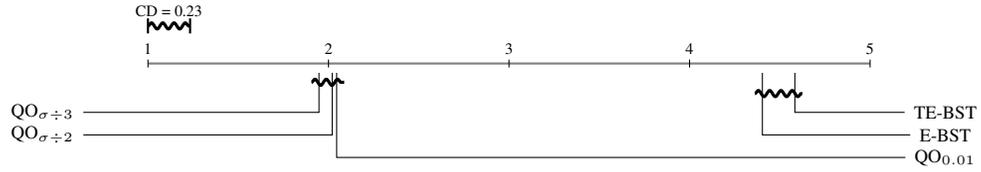

We divide the running time analysis into two parts: observation and query costs.
When observing instances, both QO variants performed better than the E-BST variants, as illustrated in the third row of Figure~\ref{fig_metrics_results}.
The differences between the QO variants were minimal in this analysis.
Interestingly, although a smaller quantization radius resulted in increased memory usage, sometimes we observed that a smaller radius resulted in faster insertions.
In other words, QO$_{\sigma \div 3}$ was faster than QO$_{\sigma \div 2}$ to monitor the incoming data.
By looking at Figure~\ref{fig_nemenyi_time_obs}, it is possible to see observe this phenomenon.
We hypothesize that this, at first glance unexpected result, comes from the fact that sometimes the addition of new elements in a hash might be faster than handling collisions.
There is, however, a delicate balance between the radius, memory usage, and running time.
As the radius decreased more, the insertions became slower since testing for membership of a hash code also has a cost.
In fact, QO$_{0.01}$ was the slowest variant of our proposal.

Surprisingly, E-BST was generally faster than its truncated counterpart.
It might be related to the same situation of handling "collisions".
We believe that, in some cases, updating the statistics of existing nodes in the E-BST is more time consuming than simply adding a new element.
Nonetheless, the differences observed between AO variants bearing from the same base algorithm were minimal.
Hence, they might not have a high impact on real-world applications.
The speed-up gains obtained by QO over E-BST (and TE-BST) were, however, clear.

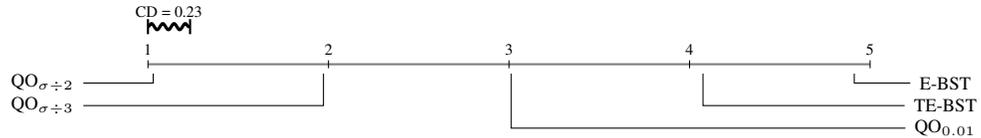
\begin{figure}[!htb]
\centering
\begin{tikzpicture}[xscale=2]
\node (Label) at (1.339943447224415, 0.7){\tiny{CD = 0.23}}; 
\draw[decorate,decoration={snake,amplitude=.4mm,segment length=1.5mm,post length=0mm},very thick, color = black] (1.2,0.5) -- (1.47988689444883,0.5);
\foreach \x in {1.2, 1.47988689444883} \draw[thick,color = black] (\x, 0.4) -- (\x, 0.6);
 
\draw[gray, thick](1.2,0) -- (6.0,0);
\foreach \x in {1.2,2.4,3.6,4.8,6.0} \draw (\x cm,1.5pt) -- (\x cm, -1.5pt);
\node (Label) at (1.2,0.2){\tiny{1}};
\node (Label) at (2.4,0.2){\tiny{2}};
\node (Label) at (3.6,0.2){\tiny{3}};
\node (Label) at (4.8,0.2){\tiny{4}};
\node (Label) at (6.0,0.2){\tiny{5}};
\node (Point) at (1.2350877192982455, 0){};\node (Label) at (0.5,-0.25){\scriptsize{QO$_{\sigma \div 2}$}}; \draw (Point) |- (Label);
\node (Point) at (2.3666666666666667, 0){};\node (Label) at (0.5,-0.55){\scriptsize{QO$_{\sigma \div 3}$}}; \draw (Point) |- (Label);
\node (Point) at (5.894736842105263, 0){};\node (Label) at (6.5,-0.25){\scriptsize{E-BST}}; \draw (Point) |- (Label);
\node (Point) at (4.889473684210527, 0){};\node (Label) at (6.5,-0.55){\scriptsize{TE-BST}}; \draw (Point) |- (Label);
\node (Point) at (3.614035087719298, 0){};\node (Label) at (6.5,-0.85){\scriptsize{QO$_{0.01}$}}; \draw (Point) |- (Label);
\end{tikzpicture}
\caption{Friedman test and Nemenyi post-hoc test when comparing the time spent by the different AO algorithms to query for split candidates ($\alpha=0.05$).}
\label{fig_nemenyi_time_query}
\end{figure}

When it comes to querying split points, the obtained results were on par with what we expected.
QO variants were clearly faster than the binary search trees.
The bottom row of Figure~\ref{fig_metrics_results} illustrates how pronounced the differences were.
Figure~\ref{fig_nemenyi_time_query} also depicts the general differences in split candidate query performance.

The QO variant with the largest radius performed faster than the other configurations, as it had fewer points to process.
When querying for split points, the higher the number of stored slots, the slower is the processing.
Finally, TE-BST was faster than its original counterpart when querying split points, as we expected.

\section{Final considerations}\label{sec_final_considerations}

This paper introduced an efficient and effective algorithm for monitoring numerical input features in online regression trees.
Our proposal, QO, requires significantly less memory and processing time than the current strategy used in practical applications.
Moreover, QO can provide accurate split point suggestions by relying on an approximate algorithm rather than a greedy approach.

The experimental results suggest that QO could be easily integrated within online regression decision tree frameworks, such as Hoeffding Trees.
In future works, we intend to evaluate the impact of using QO as attribute observers of such trees.
QO can also be easily extended to deal with multi-target regression.
We also intend to seek better alternatives to provide split candidate suggestions.
One possible strategy to follow is the usage of meta-learning to recommend the split points.
QO could be used to monitor data and provide the information necessary to induce meta-learning split point recommenders.

\section*{Acknowledgements}

The authors would like to thank FAPESP (São Paulo Research Foundation) for the financial support, by means of grant \#2018/07319-6.

\bibliographystyle{unsrt}

\end{document}